\documentclass[journal]{IEEEtran}

\usepackage{ifpdf}
\usepackage{color}
\usepackage{amssymb}
\usepackage[linesnumbered,ruled,vlined]{algorithm2e}
\usepackage{cite}
\usepackage{amsmath}
\usepackage{units}
\usepackage{array}
\usepackage{stfloats}
\usepackage{url}
\usepackage{hyperref}
\usepackage{subfigure}
\usepackage{footnote}
\usepackage{multirow}
\usepackage{booktabs}
\usepackage{tabularx}

\usepackage{orcidlink}

\begin{document}
\title{A Knowledge-Driven Diffusion Policy for End-to-End Autonomous Driving Based on Expert Routing}

\author{Chengkai~Xu$^{\orcidlink{0009-0000-6714-3004}}$,~\IEEEmembership{Student Member,~IEEE,}
        Jiaqi~Liu$^{\orcidlink{0000-0002-6171-6155}}$,~\IEEEmembership{Student Member,~IEEE,}
        Yicheng~Guo$^{\orcidlink{0009-0002-3945-2007}}$, \\
        Peng~Hang$^{\orcidlink{0000-0002-5843-0594}}$,~\IEEEmembership{Senior Member,~IEEE,}
        and~Jian~Sun$^{\orcidlink{0000-0001-5031-4938}}$ 
\thanks{
This work was supported by the National Natural Science Foundation of China (52472451, 52302502) and the Shanghai Scientific Innovation Foundation (No.23DZ1203400).
}
\thanks{
Chengkai Xu, Yicheng Guo, Peng Hang and Jian Sun are with the College of Transportation, Tongji University, Shanghai 201804, China. (e-mail: \{xuchengkai, guo\_yicheng, hangpeng, sunjian\}@tongji.edu.cn)
}
\thanks{
Jiaqi Liu is with the Department of Computer Science, University of North Carolina at Chapel Hill, United States. (e-mail:
jqliu@cs.unc.edu)
}
\thanks{Corresponding author: Peng Hang}}

\maketitle

\begin{abstract}
End-to-end autonomous driving remains constrained by the difficulty of producing adaptive, robust, and interpretable decision-making across diverse scenarios.
Existing methods often collapse diverse driving behaviors, lack long-horizon consistency, or require task-specific engineering that limits generalization.
This paper presents KDP, a knowledge-driven diffusion policy that integrates generative diffusion modeling with a sparse mixture-of-experts routing mechanism. 
The diffusion component generates temporally coherent action sequences, while the expert routing mechanism activates specialized and reusable experts according to context, enabling modular knowledge composition.
Extensive experiments across representative driving scenarios demonstrate that KDP achieves consistently higher success rates, reduced collision risk, and smoother control compared to prevailing paradigms. 
Ablation studies highlight the effectiveness of sparse expert activation and the Transformer backbone, and activation analyses reveal structured specialization and cross-scenario reuse of experts. These results establish diffusion with expert routing as a scalable and interpretable paradigm for knowledge-driven end-to-end autonomous driving. 
Validation results and demonstrations are available at
\href{https://perfectxu88.github.io/KDP-AD/}{Our Project Page}.
\end{abstract}

\begin{IEEEkeywords}
End-to-End Autonomous Vehicle; Knowledge-Driven; Diffusion Model; Mixture of Experts
\end{IEEEkeywords}

\IEEEpeerreviewmaketitle

\section{Introduction}
\IEEEPARstart 
Autonomous driving has emerged as a central challenge in intelligent transportation and machine learning research, with the promise of improving safety, efficiency, and accessibility in mobility systems \cite{zhou2024vision,zhao2024survey}. Among the various paradigms, end-to-end (E2E) learning, which directly maps raw sensory observations to driving actions, has gained significant attention for its ability to bypass hand-crafted modular pipelines and exploit the representational power of deep learning models \cite{chen2024end}.


Despite their promise, current E2E approaches remain constrained in their ability to generalize across the full diversity of real-world driving. Most methods are dominated by data-driven paradigms, whether deterministic policy regression or large-scale imitation learning. While increasing model capacity and dataset size has yielded incremental gains, this trajectory exhibits diminishing returns, which still struggle to extrapolate beyond training distributions, remain brittle in rare or unseen scenarios, and require engineering efforts that scale prohibitively with data volume \cite{cui2024survey,liu2024decision}.

Imitation learning approaches, often based on convolutional or recurrent networks, provide a direct mapping from perception to control but struggle with multi-modal distributions, typically producing averaged actions that result in unsafe or indecisive driving behaviors \cite{zhao2025survey}.
Reinforcement learning methods offer the potential for exploration and optimization under uncertainty, but remain data-hungry, unstable, and difficult to scale to real-world, safety-critical driving \cite{wu2024recent,liu2023towards}.
More recently, foundation models, such as vision-language models (VLMs) and vision-language-action models (VLAs), demonstrated remarkable abilities in reasoning, semantic grounding, and generalization. However, their application to continuous control remains limited for facing challenges in ensuring real-time inference, temporal consistency, and safety-critical reliability \cite{cui2024survey}.

In parallel, diffusion models have reshaped generative modeling in vision, audio, and control \cite{huang2025diffusion, xing2024survey}. Unlike conventional regression or classification methods, Diffusion Policy (DP) formulates action generation as a conditional denoising process, explicitly modeling multi-modal distributions, capturing sequential correlations, and providing inherently stable training \cite{chi2024diffusionpolicy, ma2024hierarchical, wen2025diffusionvla}.
However, in the context of autonomous driving, its potential has not yet been systematically explored.
By directly shaping the output action space, diffusion policies provide an expressive, robust, and temporally coherent mechanism for generating driving trajectories, making them particularly well-suited to address the core challenges of multi-modality and long-horizon stability.

Meanwhile, the Mixture of Experts (MoE) paradigm has emerged as a powerful mechanism in large-scale artificial intelligence. By sparsely activating a subset of experts, MoE enables efficient scaling, modular learning, and continual adaptation \cite{han2024fusemoe, cai2025survey}. MoE architectures have also been explored in autonomous driving, for example in multitask policies and modular prediction frameworks \cite{yang2025drivemoe}.
However, such uses often treat experts as task-specific modules, limiting their transferability and failing to exploit their potential as reusable components of driving knowledge.


To overcome these limitations, we introduce KDP, a \textbf{K}nowledge-Driven \textbf{D}iffusion \textbf{P}olicy for end-to-end autonomous driving. KDP integrates diffusion-based generative modeling with sparse expert routing. The diffusion component offers robust distributional expressivity for action generation, while the MoE mechanism organizes experts as abstract driving knowledge units. Each expert encodes structured capabilities—such as longitudinal regulation, interaction handling, or lateral negotiation—that can be selectively composed according to scenario demands. This knowledge-driven framework enables expert reuse across scenarios and supports the emergence of new behaviors through compositionality.
Experiments demonstrate that our approach establishes a modular, generative, and knowledge-driven framework for autonomous driving that enhances adaptability and generalization while supporting compositional reuse of knowledge.

The main contributions of this paper are as follows:

\begin{itemize}
    \item A knowledge-driven end-to-end driving framework.
    We remodel experts in MoE as abstract driving knowledge units, enabling modular and compositional policy learning beyond task-centric formulations.
    \item  Integration of diffusion modeling with expert routing. By combining diffusion policies with MoE, KDP achieves expressive action generation together with modular adaptability across diverse scenarios.
    \item Dynamic composition and continual adaptability. Sparse activation is integrated into the framework, enabling the model to dynamically compose driving behaviors by combining reusable knowledge units, which facilitates long-term adaptability and transferability across multiple driving environments.
\end{itemize}

\section{Related Works}
\subsection{End-to-End Autonomous Driving}
\begin{figure}
    \centering
    \includegraphics[width=\linewidth]{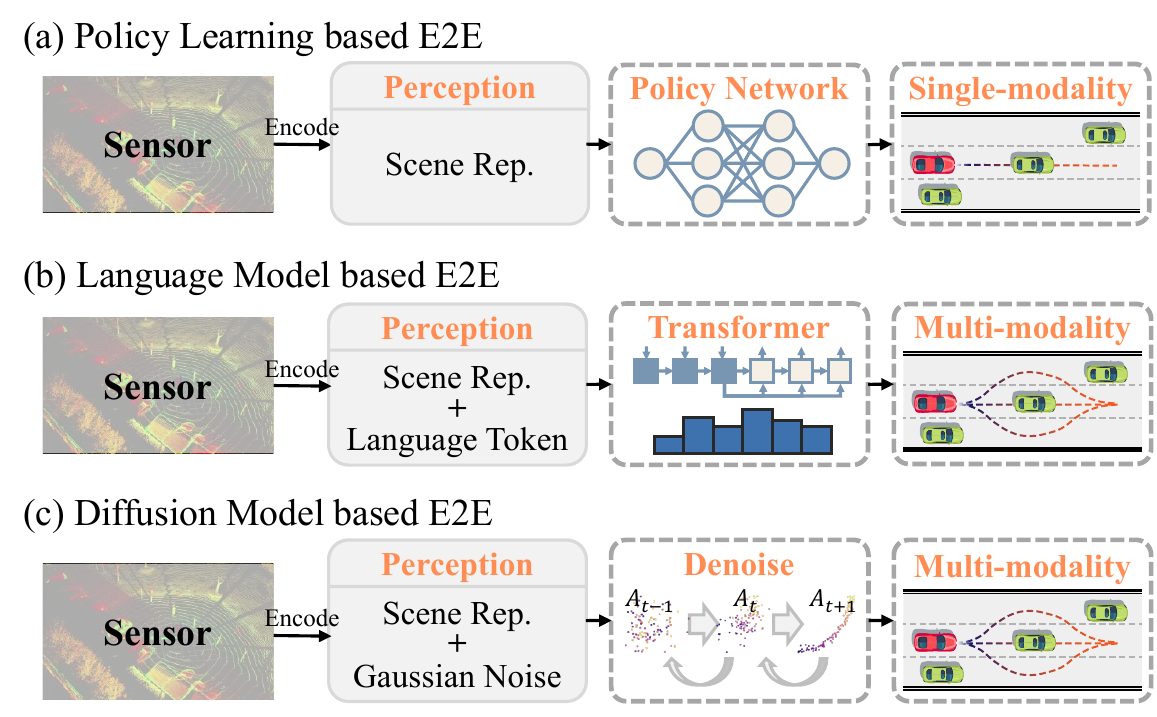}
    \caption{
    End-to-end autonomous driving paradigms: (a) policy networks, (b) language-based model, and (c) diffusion policies, where single-modality refers to policies that collapse multiple feasible maneuvers into a single trajectory, whereas multi-modality denotes the ability to generate diverse valid trajectories under the same driving context.
    }
    \label{fig:e2e}
\end{figure}

Early E2E approaches to autonomous driving sought to directly map sensory observations to control commands through imitation learning, particularly convolutional network–based models \cite{zheng2022imitation, jiang2023vad} shown in Fig.~\ref{fig:e2e} (a). While effective in simple scenarios, these deterministic architectures collapse multi-modal driving behaviors into averaged predictions, leading to indecisive or unsafe actions. Reinforcement learning methods have also been explored, offering the potential to optimize long-term objectives under uncertainty, but they suffer from excessive data requirements, unstable training dynamics, and limited scalability in real-world safety-critical environments \cite{xu2025tell, huang2021deductive}.

To address these limitations, recent work has adopted sequence modeling with transformers to better capture temporal dependencies and fuse multi-modal inputs, which unifies perception, prediction, and planning within a transformer-based framework \cite{shao2023safety, hu2023planning}. As illustrated in Fig.~\ref{fig:e2e} (b), these models demonstrate improved integration of spatiotemporal information and better scene understanding. However, they remain fundamentally deterministic while generating single future trajectories without accounting for the inherent multi-modality of driving decisions, and they struggle to guarantee long-horizon stability in action sequences.

More recently, researchers have begun to investigate the role of generative AI in driving. VLMs and VLAs have been explored for semantic grounding and reasoning about traffic scenarios \cite{shao2024lmdrive, nie2024reason2drive}. These approaches highlight the promise of generative reasoning for interpretability and generalization. However, their practical use in closed-loop driving remains limited due to high latency, lack of temporal consistency in control, and challenges in producing reliable continuous actions. In this context, diffusion policy, which is shown in Fig.~\ref{fig:e2e} (c), provides a principled generative mechanism to address these challenges, which can explicitly model multi-modality and ensure temporal stability \cite{yang2023diffusion}.

\subsection{Diffusion Models for Autonomous Systems}
Diffusion models have recently emerged as a powerful class of generative methods, offering stable training, likelihood-based optimization, and strong capacity to model multi-modal distributions \cite{huang2025diffusion, cao2024survey}. Unlike variational autoencoders (VAEs) or generative adversarial networks (GANs), diffusion models avoid issues such as posterior collapse or adversarial instability and consistently yield diverse and high-fidelity samples \cite{he2025diffusion, tang2023emergent}. These properties make diffusion models particularly appealing for sequential decision-making in embodied agents, where policies must capture uncertainty, generate temporally consistent actions, and remain robust under complex input distributions.

Building on these advances, DP has been introduced as a framework for robotic control. In this formulation, action trajectories are generated through a conditional denoising process, allowing the policy to represent multi-modal behaviors while maintaining temporal coherence over extended horizons \cite{chi2024diffusionpolicy, li2024learning}. Empirical results demonstrate that DP achieves superior performance in manipulation and locomotion tasks, outperforming conventional imitation learning and reinforcement learning baselines in both stability and robustness \cite{ma2024hierarchical}. By treating trajectory generation as a generative modeling problem, DP offers a principled solution to the fundamental challenges of multi-modality, sequential correlation, and uncertainty in embodied control \cite{wen2025diffusionvla, carvalho2025motion}.

Building on these advances, recent studies have begun adapting diffusion models to autonomous driving, exploring strategies such as truncated diffusion schedules, cascade decoders, and classifier guidance to improve efficiency, diversity, and safety in trajectory generation \cite{liao2025diffusiondrive, zheng2025diffusion}.
While these efforts demonstrate the promise of diffusion for planning, they largely retain a monolithic generation paradigm that optimizes sampling efficiency but does not address the broader challenges of modularity, adaptability, and knowledge structuring \cite{jiang2025diffvla, wang2025diffad}. 
This limitation motivates a more general framework that couples diffusion modeling with mechanisms for expert specialization and reuse, enabling multi-modal and temporally stable decision generation while supporting continual learning and transferable driving knowledge.

\begin{figure*}
    \centering
    \includegraphics[width=\linewidth]{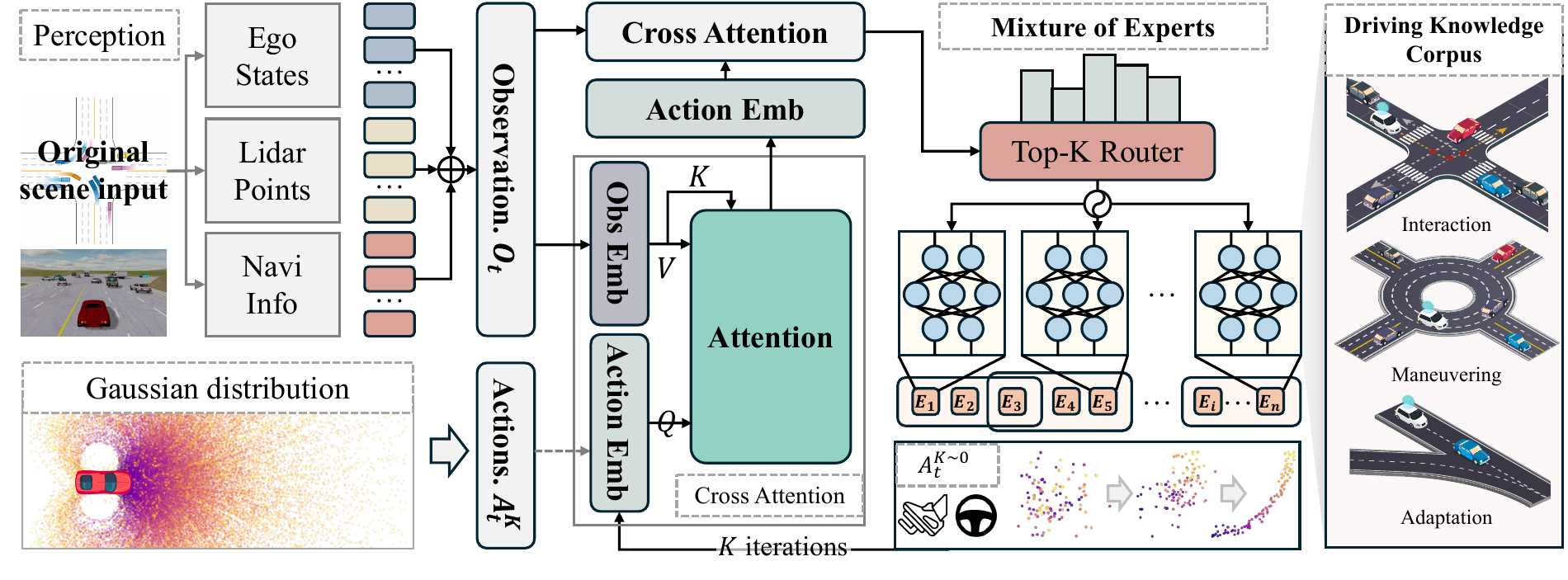}
    \caption{Framework of the proposed Knowledge-Driven Diffusion Policy. Scene inputs condition a diffusion-based policy to generate multi-modal, temporally coherent action sequences.
    A Mixture-of-Experts module refines these sequences by activating experts interpreted as abstract knowledge units, whose combinations express diverse and extensible driving skills such as interaction, maneuvering, and adaptation.
    }
    \label{fig:framework}
\end{figure*}

\subsection{Mixture-of-Experts in Generative AI}
The MoE paradigm was originally proposed as a way to decompose complex functions into multiple specialized submodels, with a gating mechanism dynamically selecting which experts to activate. In modern deep learning, MoE has emerged as a central architecture for scaling models efficiently. By sparsely activating only a small subset of experts per input, MoE combines the benefits of massive model capacity with manageable computational cost \cite{wu2024multi, ding2025denseformer}. This design enables modular specialization, where experts capture different aspects of the input distribution, while maintaining inference efficiency.

MoE architectures have achieved remarkable success in generative artificial intelligence, particularly in large-scale language and vision models \cite{cai2025survey, riquelme2021scaling}. In natural language processing, models such as DeepSeek-v3 and DeepSeek-R1 \cite{liu2024deepseek, guo2025deepseek} demonstrate that MoE can scale to trillions of parameters while retaining efficiency, with experts specializing in different linguistic or semantic patterns. In vision and multi-modal learning, MoE has been incorporated into generative tasks such as image synthesis and text-to-image diffusion, enabling different experts to capture distinct modalities or generative styles. These advances highlight MoE as a general mechanism for modularity and diversity in generative modeling.

Beyond language and vision, MoE has also been investigated in embodied intelligence, including robotics and autonomous driving. In robotics, MoE modules have been applied to multitask learning and skill decomposition, improving transferability and continual learning \cite{chen2023mod, song2024germ}. In autonomous driving, MoE has been explored for multi-task prediction and decision-making frameworks, but primarily in a task-centric manner rather than within a generative policy formulation \cite{liao2025diffusiondrive, wan2025geminus}. Thus, despite its promise, the use of MoE for structuring abstract driving knowledge in generative decision-making remains underexplored. This gap motivates the integration of MoE into diffusion policy to realize a knowledge-driven framework for end-to-end autonomous driving.

\section{Methodology}

\subsection{Framework Overview}
The overall framework of the proposed method is illustrated in Fig.~\ref{fig:framework}. Following an end-to-end paradigm, the system directly maps driving observations to control actions through a generative decision-making process. The inputs consist of heterogeneous scene information, including ego states, LiDAR point clouds, and high-level navigation signals, providing a comprehensive representation of the driving environment that conditions subsequent policy generation.

At the core lies a diffusion-based policy backbone, where action trajectories are sampled via a conditional denoising process. Starting from Gaussian noise, trajectories are progressively refined under the guidance of scene observations. Through attention-mediated interactions between observation and action embeddings, the model captures temporal dependencies and generates smooth, coherent long-horizon action sequence. This formulation naturally accommodates the multi-modality of driving behavior, allowing the policy to represent multiple feasible maneuvers within the same context.

To enhance modularity and adaptability, a MoE module is integrated. A Top-K router dynamically activates a subset of experts according to the driving scenario, and the selected experts collaboratively refine the action predictions. Each expert can be interpreted as an abstract knowledge component, with different expert combinations encoding distinct driving competencies such as negotiating intersections, navigating roundabouts, or adapting to unfamiliar road structures. By coupling generative policy learning with structured knowledge representations, the MoE design enables the framework to achieve both temporal stability and continual adaptability in complex, dynamic traffic environments.

\subsection{Problem Formulation}
\label{Problem_Formulation}
We formulate end-to-end autonomous driving as a sequential decision-making problem, where the objective is to learn a policy that maps sensory observations to continuous control actions. At each time step $t$, the system receives an observation $o_t \in \mathcal{O}$, which in our setting comprises the ego-vehicle state, LiDAR point clouds, and high-level navigation commands. The policy generates an action $a_t \in \mathcal{A}$, defined as a pair of low-level control signals including throttle and steering angle.

Rather than predicting single-step actions independently, the policy generates temporally coherent action sequences. Over a planning horizon $H$, the sequence is denoted as

\begin{equation}
    a_{t:t+H-1} = \{a_t, a_{t+1}, \ldots, a_{t+H-1}\}, \quad a_\tau \in \mathcal{A}
\end{equation}
which characterizes the evolution of vehicle behavior across multiple future steps. The learning objective is therefore to model the conditional distribution of action sequences given observation:

\begin{equation}
\label{pi_theta}
    \pi_\theta(a_{t:t+H-1} \mid o_{t})
\end{equation}
where $\pi_\theta$ is parameterized by a neural policy with parameters $\theta$. This formulation naturally accounts for both the multi-modality of feasible maneuvers and the requirement of temporal consistency across long horizons.

In practice, training data are obtained from expert demonstrations in the form of trajectories

\begin{equation}
    \tau = \{(o_1,a_1), (o_2,a_2), \ldots, (o_T,a_T)\}
\end{equation}
where each pair $(o_t, a_t)$ represents an observation and its corresponding ground-truth control signal. The learning task is to approximate the expert’s distribution over action sequences by maximizing the likelihood of demonstrated behavior:

\begin{equation}
    \max_\theta \; \mathbb{E}_{\tau \sim \mathcal{D}} \big[ \log \pi_\theta(a_{t:t+H-1} \mid o_{t}) \big]
\end{equation}
with $\mathcal{D}$ denoting the dataset of demonstrations.

Through this formulation, end-to-end autonomous driving is cast as conditional sequence modeling. Given past and current observations, the policy generate a distribution over plausible future action sequences that align with expert demonstrations and ensure safe, coherent vehicle control.

\subsection{Diffusion Policy for Driving}

\begin{figure}
    \centering
    \includegraphics[width=\linewidth]{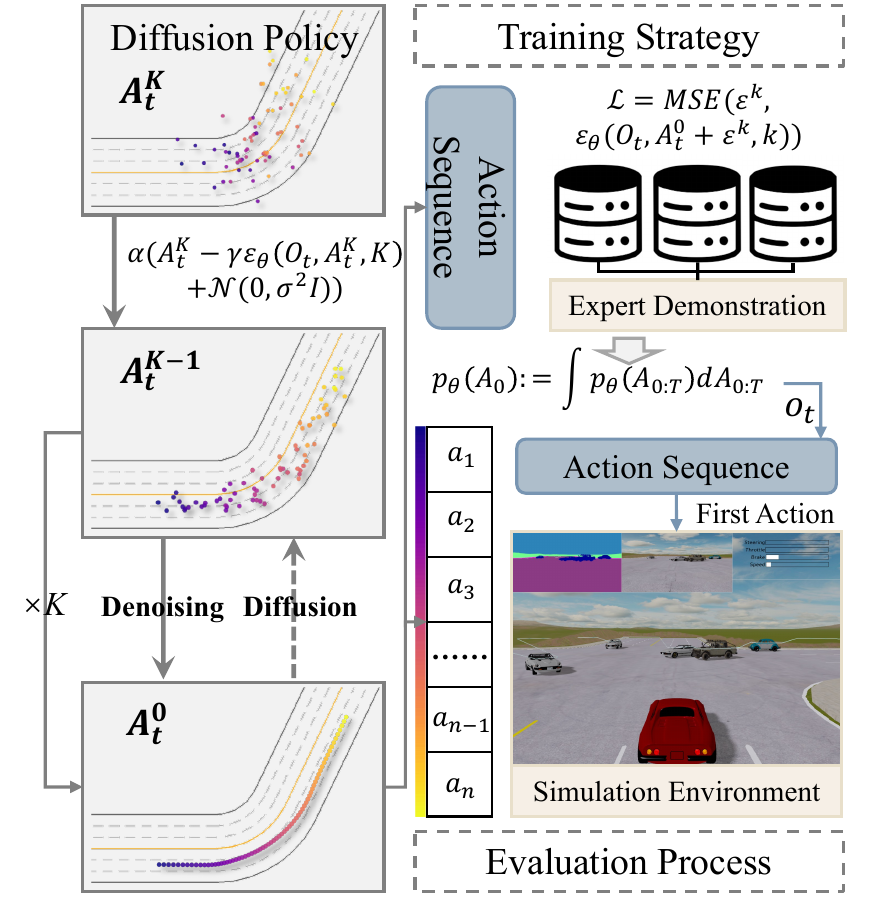}
    \caption{Diffusion-based action generation, where expert demonstrations are perturbed by forward diffusion and recovered by reverse denoising to produce executable action sequences for simulation.}
    \label{fig:diffusion_policy}
\end{figure}

\subsubsection{\textbf{Generative Policy Formulation}}

We adopt a generative modeling perspective for end-to-end driving, where the policy is treated as a probabilistic model that generates diverse and temporally coherent action sequences conditioned on the driving context instead of directly regressing control commands from observations. This approach is particularly suitable for autonomous driving, where multiple maneuvers may be equally valid under identical conditions, such as braking to yield or changing lanes. A deterministic mapping tends to collapse such alternatives into averaged actions, which may be indecisive or unsafe.

Formally, the generative policy can be interpreted as learning a latent-variable model:

\begin{equation}
    a_{t:t+H-1} = f_\theta(z, o_{t}), \quad z \sim p(z)
\end{equation}
where $a_{t:t+H-1}$ is the sequence of throttle–steering actions over horizon $H$, $o_t$ is the conditioning observation, and $z$ is a latent variable drawn from a simple prior such as Gaussian noise. Policy learning thus reduces to designing an effective transformation that maps noise samples into realistic action sequences consistent with expert demonstrations. 

This perspective naturally motivates the adoption of diffusion models. The overall process of diffusion-based action generation is illustrated in Fig.~\ref{fig:diffusion_policy}. Expert demonstrations provide ground-truth action sequences, which are progressively corrupted through the forward diffusion process and subsequently reconstructed by the reverse denoising policy. During inference, the policy iteratively refines Gaussian noise into executable action sequences that are evaluated within the simulation environment.

\subsubsection{\textbf{Forward Diffusion of Action Sequences}}
The forward diffusion process defines how clean action sequences from expert demonstrations are progressively perturbed with Gaussian noise to construct training targets. Let an expert action sequence over horizon $H$ be denoted as $a_{0} \in \mathbb{R}^{H \times d}$, where $d$ is the action dimension. Following the denoising diffusion probabilistic model (DDPM)\cite{ho2020denoising}, we define a Markov chain that gradually corrupts $a_{0}$ into a sequence of increasingly noisy representations ${a_{1}, \ldots, a_{T}}$:

\begin{equation}
    q(a_{t} \mid a_{t-1}) = \mathcal{N}\big(\sqrt{1-\beta_t} \, a_{t-1}, \, \beta_t I \big), \quad t = 1, \ldots, T
\end{equation}
where $\{\beta_t\}_{t=1}^T$ is a variance schedule. Iterating this process transforms $a_0$ into nearly isotropic Gaussian noise $a_T$.

A key property of the diffusion process is that $a_t$ can be expressed in the closed form relative to the clean sequence $a_0$:

\begin{equation}
    q(a_t \mid a_0) = \mathcal{N}\big(\sqrt{\bar{\alpha}_t} \, a_0, \, (1-\bar{\alpha}_t) I \big)
\end{equation}
where $\alpha_t = 1 - \beta_t$ and $\bar{\alpha}_t = \prod_{s=1}^t \alpha_s$. This formulation allows the training procedure to sample $a_t$ at any diffusion step directly, without simulating the entire Markov chain.

This forward process can be interpreted as perturbing expert action sequences into progressively noisier versions while retaining statistical dependence on the ground-truth actions. These noisy samples serve as supervision targets for the reverse denoising process, where the policy learns to iteratively recover clean, temporally coherent action sequences from corrupted inputs conditioned on scene observations.

\begin{algorithm}
    \caption{Training of KDP}
    \label{alg:kdp_train}
    \SetAlgoLined
    \SetKwInOut{Input}{Input}\SetKwInOut{Output}{Output}
    
    \Input{Demonstrations $\mathcal{D}$; diffusion steps $T$; noise schedule $\{\beta_t\}_{t=1}^{T}$; Top-$K$}
    \Output{Trained parameters $\Theta$}
    
    \BlankLine
    \textbf{Precompute:} \\
    $\alpha_t \leftarrow 1-\beta_t,\;\; \bar\alpha_t \leftarrow \prod_{s=1}^{t}\alpha_s$ for $t{=}1{:}T$
    
    \BlankLine
    \For{\textnormal{training iterations}}{
        \textbf{Sample minibatch:} \\
        \quad Sample trajectory $\tau{\sim}\mathcal{D}$ \\
        \quad Sample $t {\sim} \mathrm{Uniform}\{1,\ldots,T\}$ \\
        \quad Build context $c_t \leftarrow \mathrm{Enc}(o_{t})$ \\
        \quad Extract aligned clean sequence $a_0 \in \mathbb{R}^{H\times d}$

        \BlankLine
        \textbf{Forward diffusion:} \\ 
        \quad Sample $\epsilon {\sim} \mathcal{N}(0,I)$ \\
        \quad Set $a_t \leftarrow \sqrt{\bar\alpha_t}\,a_0 \;+\; \sqrt{1-\bar\alpha_t}\,\epsilon$

        \BlankLine
        \textbf{MoE noise prediction:} \\
        \quad $x \leftarrow \mathrm{Emb}(a_t,c_t,t)$ \\
        \quad $\mathbf{g} \leftarrow \mathrm{softmax}(xW)$, $\mathcal{S} \leftarrow \mathrm{Top}\text{-}K(\mathbf{g})$ \\
        \quad $\tilde{\mathbf{g}} \leftarrow \mathrm{Renorm}(\mathbf{g}\!\restriction\!\mathcal{S})$ \\
        \quad $\hat\epsilon \leftarrow \sum_{i \in \mathcal{S}} \tilde g_i \, E_i(a_t,c_t,t)$

        \BlankLine
        \textbf{Losses:}\\
        \quad Denoising loss: $L_{\text{diff}} \leftarrow \|\epsilon - \hat\epsilon\|^2$ \\
        \quad Load balancing:\\
        \quad \quad $p_i \leftarrow \frac{1}{B}\sum_{b=1}^{B}\mathbb{I}[i\!\in\!\mathcal{S}(x^{(b)})]$ \\
        \quad \quad $L_{\text{bal}} \leftarrow -\tfrac{1}{N}\sum_i p_i \log p_i$ \\

        \quad Compute empirical joint $\hat p(K,E)$ \\
        \quad $L_{\text{mi}} \leftarrow - \sum_{j,i} \hat p(K_j,E_i)\log\frac{\hat p(K_j,E_i)}{\hat p(K_j)\hat p(E_i)}$
        
        \textbf{Total loss and Update:} \\
        \quad $L \leftarrow L_{\text{diff}} + \lambda_{\text{bal}} L_{\text{bal}} + \gamma_{\text{mi}} L_{\text{mi}}$ \\
        \quad $\Theta \leftarrow \Theta - \eta \,\nabla_\Theta L$
    }
\end{algorithm}

\subsubsection{\textbf{Reverse Denoising Process as Driving Policy}}
The driving policy is parameterized as the reverse denoising process that transforms noise samples into executable action sequences conditioned on observations. Starting from an initial sample $a_T \sim \mathcal{N}(0, I)$, the policy progressively refines this noise through $T$ denoising steps until a clean action sequence $a_0$ is recovered. The reverse process is defined as a Markov chain with learnable Gaussian transitions:

\begin{equation}
\begin{aligned}
p_\theta(a_{t-1} \mid a_t, o_{t})
&= \mathcal{N}\!\big(
      \mu_\theta(a_t, o_{t}, t), \\
&\quad \ \ \Sigma_\theta(a_t, o_{t}, t)
   \big), \quad t = T,\ldots,1
\end{aligned}
\end{equation}
where $\mu_\theta$ and $\Sigma_\theta$ are outputs of a neural network parameterized by $\theta$ and conditioned on $o_t$.

Following DDPM and Diffusion Policy, the network is trained to predict the injected noise $\epsilon$ rather than reconstruct $a_0$ directly. 

At inference time, the learned policy iteratively applies the reverse process to sample action sequences:

\begin{equation}
\begin{aligned}
a_{t-1} 
&= \frac{1}{\sqrt{\alpha_t}} 
   \Big( a_t 
      - \frac{1-\alpha_t}{\sqrt{1-\bar{\alpha}_t}}
        \, \epsilon_\theta(a_t, o_{t}, t) \Big) \\
&\quad + \sigma_t z, 
\qquad z \sim \mathcal{N}(0,I)
\end{aligned}
\end{equation}
where $\alpha_t$ and $\sigma_t$ are defined by the noise schedule. This iterative refinement ensures that the sampled sequences remain consistent with the conditional distribution $p(a_0 \mid o_t)$.

This reverse denoising process constitutes the driving policy where observations condition the generative dynamics, and each sampled sequence represents a coherent, multi-modal driving maneuver. By explicitly modeling the distribution over feasible action sequences, the policy captures the variability of human demonstrations while ensuring temporal stability across the planning horizon.

\subsubsection{\textbf{Training Objective}}
The training objective of the diffusion-based driving policy is to align the reverse denoising process with the expert action distribution. Given a clean expert sequence $a_0$ and a random diffusion step $t \sim \text{Uniform}\{1,\ldots,T\}$, a noisy sequence is constructed as

\begin{equation}
\label{eq:noisy_sequence}
    a_t = \sqrt{\bar{\alpha}_t}\, a_0 + \sqrt{1-\bar{\alpha}_t}\, \epsilon, \quad \epsilon \sim \mathcal{N}(0,I)
\end{equation}

The network $\epsilon_\theta$ is then trained to predict $\epsilon$ from $(a_t, o_t, t)$. The loss function is defined as:

\begin{equation}
    \label{eq:denoising_loss}
    \mathcal{L}_{\text{denoise}}(\theta) = 
        \mathbb{E}_{a_0, \epsilon, t}\!\left[
            \|\epsilon - \epsilon_\theta(a_t, o_t, t)\|^2
        \right].
\end{equation}

Minimizing this objective is equivalent to maximizing a variational lower bound on the likelihood of expert demonstrations, thereby ensuring that the learned policy approximates the true conditional distribution of action sequences $\pi(a_{t\:t+H-1} \mid o_{t})$.  The complete KDP training procedure, which integrates loss denoising and routing constraints, is summarized in Algorithm~\ref{alg:kdp_train}.

This denoising loss provides the core training signal for diffusion-based action generation. In our framework, additional terms related to expert routing are introduced when integrating the MoE module, as described in Section~\ref{sec:MoE}. This objective encourages the policy to generate action sequences that remain close to expert trajectories under perturbations, while retaining the multi-modal flexibility afforded by the generative formulation.

\subsection{MoE-based Knowledge Routing}
\label{sec:MoE}

\begin{figure}
    \centering
    \includegraphics[width=0.9\linewidth]{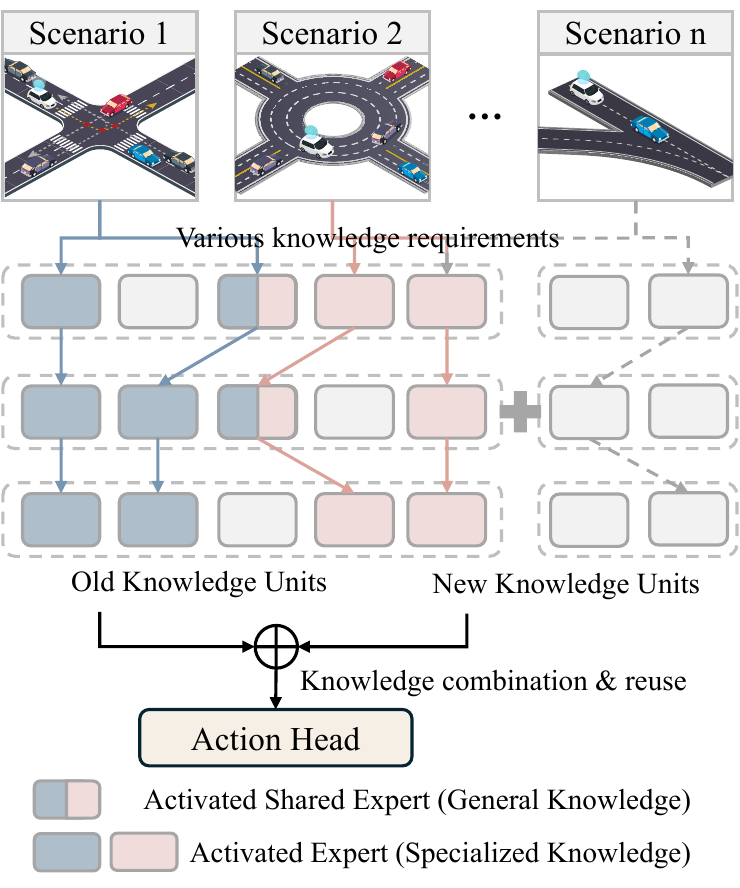}
    \caption{MoE-based knowledge routing framework, where a Top-$K$ router dynamically selects specialized experts to compose modular driving knowledge for adaptive policy learning.}
    \label{fig:moe}
\end{figure}

\subsubsection{\textbf{Integration of MoE into Diffusion Policy}}

To extend the diffusion-based driving policy with modularity and scalability, a MoE mechanism as illustrated in Fig.~\ref{fig:moe} is incorporated into the denoising network. In this design, the policy is no longer represented by a single monolithic model but instead by a collection of experts, each responsible for predicting denoising signals under specific contexts. Formally, we define a set of experts

\begin{equation}
    \{E_i\}_{i=1}^N, \quad E_i: \mathbb{R}^{H \times d} \times \mathcal{O} \times \{1,\ldots,T\} \to \mathbb{R}^{H \times d}
\end{equation}
where each expert $E_i$ is parameterized by $\theta_i$ and outputs a noise estimate $\epsilon_{\theta_i}(a_t, o_t, t)$ given a noisy action sequence $a_t$, the conditioning observation $o_t$, and the diffusion timestep $t$.

Expert selection is governed by a gating network $g_\phi(o_t)$, parameterized by $\phi$, which produces a probability distribution over experts based on the current observation. To ensure computational efficiency, we adopt a sparse routing strategy in which only the top-$K$ experts with the highest gate scores are activated. Denoting the active expert set at time $t$ as $\mathcal{S}_t \subset \{1, \ldots, N\}$, the aggregated noise prediction is expressed as

\begin{equation}
    \epsilon_\theta(a_t, o_t, t) = \sum_{i \in \mathcal{S}_t} g_\phi^i(o_t) \, \epsilon_{\theta_i}(a_t, o_t, t)
\end{equation}
where $g_\phi^i(o_t)$ denotes the normalized routing weight assigned to expert $E_i$.

This formulation allows the diffusion policy to dynamically compose multiple experts in response to changing driving contexts. The sparse gating mechanism ensures that, although the full model has large capacity through multiple experts, only a small subset contributes to each forward pass, thus retaining computational efficiency. By integrating MoE into the diffusion policy in this manner, the generative process is enhanced with structural modularity with experts providing fine-grained modeling capacity and the router orchestrating their combination according to scene observations.

\subsubsection{\textbf{Routing Mechanism}}
In the integrated MoE–diffusion policy, expert selection is governed by a routing network that assigns probabilities over the available experts based on the current observation. Formally, let $x \in \mathbb{R}^m$ denote the input representation derived from the observation–action embedding at a given diffusion step. The router computes a score vector $s = xW \in \mathbb{R}^N$, where $W \in \mathbb{R}^{m \times N}$ is a learnable weight matrix and $N$ is the number of experts. The scores are normalized with a softmax function to obtain a routing distribution

\begin{equation}
    g_\phi(x) = \text{softmax}(s)
\end{equation}

To maintain efficiency, only the top-$K$ experts are activated for each input. Let $\mathcal{S}(x)$ denote the indices of the $K$ largest elements in $g_\phi(x)$. The contribution of expert $E_i$ is then masked if $i \notin \mathcal{S}(x)$, resulting in a sparse aggregation:

\begin{equation}
    y(x) = \sum_{i \in \mathcal{S}(x)} g_\phi^i(x) \, E_i(x)
\end{equation}
where $g_\phi^i(x)$ is the normalized routing weight of expert $E_i$. This sparse gating strategy ensures that only a small fraction of experts contribute to each forward pass, while the inactive experts remain computationally dormant.

The use of top-$K$ routing promotes specialization and efficiency and plays a central role in dynamically composing driving knowledge. By conditioning on the embedded representation of LiDAR, ego state, and navigation information, the router selects those experts most relevant to the current traffic scenario. At the same time, the sparsity constraint ensures that the active parameter count remains nearly constant regardless of the total number of experts, which allows the policy to scale in capacity without proportional increases in computation.

\subsubsection{\textbf{Training Objective with Routing Constraints}}

The training objective of the integrated MoE–diffusion policy extends the standard denoising loss with additional constraints to regulate expert utilization. Directly minimizing the diffusion denoising loss often causes the router to favor a small subset of experts, leading to expert overload and limited specialization. To mitigate this, auxiliary regularization terms are incorporated to promote balanced and structured expert usage.

Formally, given an expert action sequence $a_0$, a noisy sequence is constructed as Eq.~\ref{eq:noisy_sequence}, and the denoising loss is defined as Eq.~\ref{eq:denoising_loss}. To avoid router collapse and ensure balanced allocation of experts, we add a load-balancing term.  Let $p_i$ denote the probability that expert $E_i$ is selected within a minibatch. The balancing objective is given by

\begin{equation}
    \mathcal{L}_{\text{balance}} = \frac{1}{N} \sum_{i=1}^N \left( p_i \log p_i \right)
\end{equation}
which penalizes skewed utilization by increasing the entropy of expert selection. In addition, a mutual information regularizer is adopted to specialize in distinct knowledge units. Let $K$ denote the space of abstract knowledge categories, then the mutual information between experts and knowledge components is maximized:

\begin{equation}
    I(K, E) = \sum_{j=1}^{J} \sum_{i=1}^{N} p(K_j,E_i) \log \frac{p(K_j,E_i)}{p(K_j)\,p(E_i)}
\end{equation}
where $p(K_j)$ is assumed uniform and $p(K_j,E_i)$ is estimated from expert selection statistics. By maximizing $I(K,E)$, the training process encourages experts to align with distinct knowledge representations, ensuring that the ensemble of experts covers a diverse corpus of driving knowledge. The composition of these experts, as selected by the router, then implicitly corresponds to appropriate behaviors in particular scenarios.

The overall training objective is therefore expressed as

\begin{equation}
    \mathcal{L} = \mathcal{L}_{\text{denoise}} 
    + \lambda_{\text{bal}} \mathcal{L}_{\text{balance}} 
    - \gamma I(K,E)
\end{equation}
where $\lambda_{\text{bal}}$ and $\gamma$ control the strength of the regularization terms. At deployment time, the policy samples executable action sequences via iterative denoising, as outlined in Algorithm~\ref{alg:kdp_infer}.

This formulation ensures that the policy not only learns to denoise action sequences effectively but also develops a structured allocation of expertise grounded in knowledge. The denoising loss aligns the generative process with expert demonstrations, while the routing constraints encourage diversity, balance, and the emergence of specialized knowledge units that can be flexibly composed to handle diverse driving situations.

\begin{figure}
    \centering
    \includegraphics[width=\linewidth]{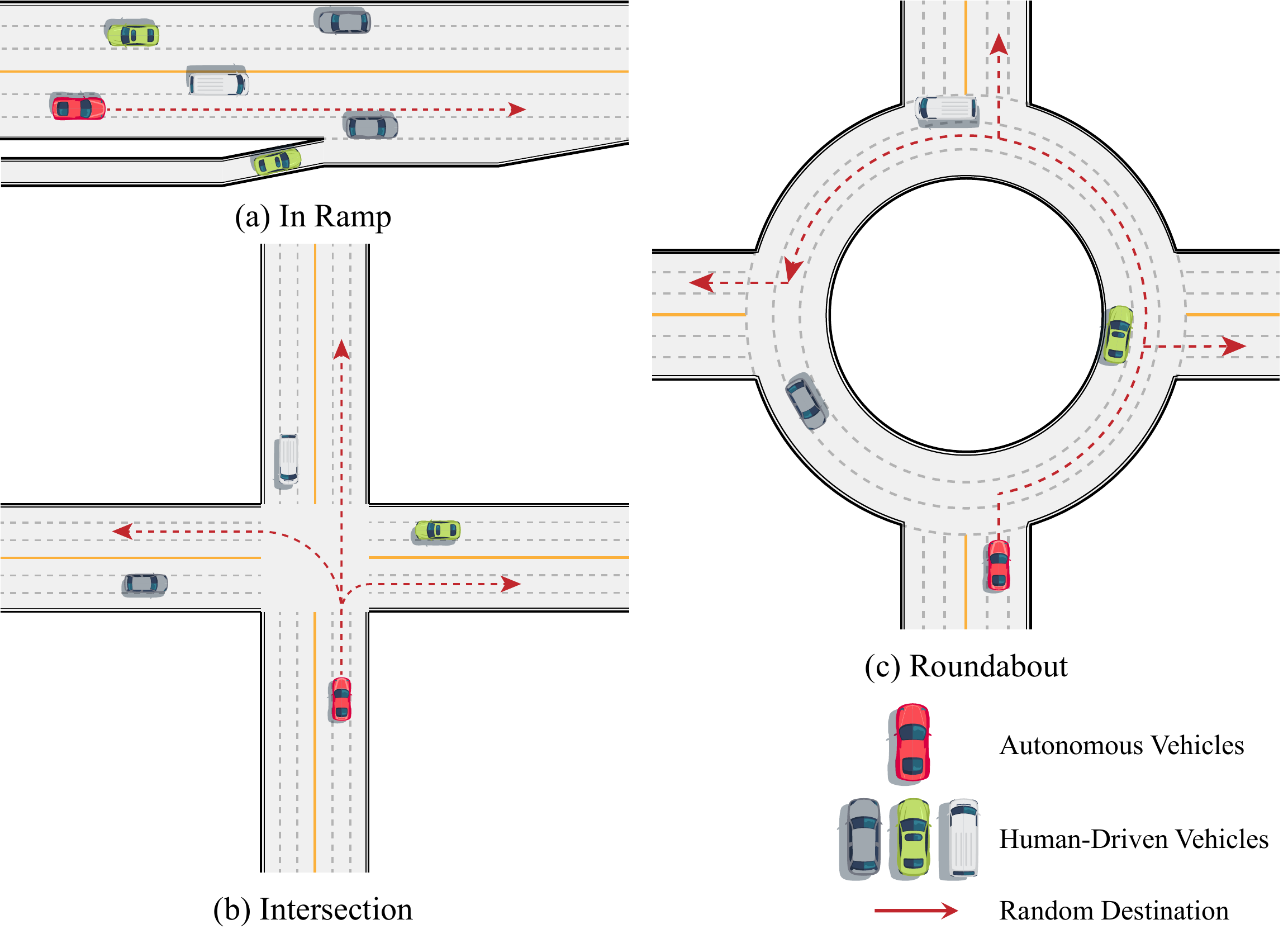}
    \caption{Three selected representative and challenging scenarios , including (a) In Ramp, (b) Intersection, and (c) Roundabout, which increase in difficulty and test different aspects of end-to-end autonomous driving capabilities.}
    \label{fig:scenario}
\end{figure}

\begin{algorithm}
    \caption{Online Inference of KDP}
    \label{alg:kdp_infer}
    \SetAlgoLined
    \SetKwInOut{Input}{Input}\SetKwInOut{Output}{Output}
    \Input{Current observations $o_{t}$, trained $\Theta$, schedule $\{\alpha_t,\bar\alpha_t\}_{t=1}^{T}$, Top-$K$, stochasticity $\eta\!\in\![0,1]$}
    \Output{Executable action sequence $\hat a_{0}\in\mathbb{R}^{H\times d}$}
    \BlankLine
    $c_t \leftarrow \mathrm{Enc}(o_{t})$;\quad $a_T \sim \mathcal{N}(0,I)$ \\
    \For{$t \leftarrow T$ \KwTo $1$}{
        \textbf{MoE noise prediction:} 
        $\hat\epsilon \leftarrow \epsilon_\Theta(a_t,c_t,t)=\sum_{i\in\mathcal{S}(x)} \tilde g_i(x)\,E_i(a_t,c_t,t)$ \\
        \textbf{Mean update:}
        $\mu_t \leftarrow \frac{1}{\sqrt{\alpha_t}}\!\left(a_t - \frac{1-\alpha_t}{\sqrt{1-\bar\alpha_t}}\,\hat\epsilon\right)$\\
        \textbf{Stochasticity:}
        \begin{small}
        $\sigma_t \leftarrow \eta\, \sqrt{\frac{1-\bar\alpha_{t-1}}{1-\bar\alpha_t}}\,\sqrt{1-\alpha_t}$
        \end{small}\\
        \uIf{$t>1$}{Sample $z \sim \mathcal{N}(0,I)$}
        \Else{$z \leftarrow 0$}
        $a_{t-1} \leftarrow \mu_t + \sigma_t z$
    }
    \textbf{return} $\hat a_{0} \leftarrow a_{0}$
\end{algorithm}

\section{Experiment and Evaluation}

\subsection{Experimental Setup}
We evaluate our method in a simulation environment built on MetaDrive \cite{li2022metadrive}, which supports flexible configuration of road geometries and traffic flows. As shown in Fig.~\ref{fig:scenario}, three representative scenarios are selected, arranged from lower to higher difficulty: In Ramp, Intersection, and Roundabout. The In Ramp scenario consists of a 150 m main road with a 50 m merging segment, testing longitudinal negotiation during cut-ins. The Intersection is modeled as an unsignalized six-lane crossing with a 50 m interaction zone, while the Roundabout comprises three lanes with a 50 m entry section and a 70 m outer diameter, requiring complex multi-agent navigation. To maximize interaction, traffic vehicles are governed by IDM and MOBIL with randomized parameters and are triggered dynamically as the ego vehicle approaches.

The policy receives as input a unified observation vector comprising (i) the ego state, including speed, steering, heading deviation, and recent control history, (ii) high-level navigation signals projected into the ego coordinate frame, and (iii) 240 LiDAR points sampled over a 360° field of view with a 50 m perceptual range. The action space is a normalized two-dimensional vector $a = [a_1,a_2] \in [-1,1]^2$, which is mapped to steering, acceleration, and braking commands. The steering command $u_s$ is expressed in degrees, the engine force $u_a$ in horsepower, and the braking force $u_b$ in horsepower, which are derived as

\begin{equation}
\begin{cases}
    u_s = S_{\max} \, a_1 \\
    u_a = F_{\max} \, \max(0,a_2) \\
    u_b = B_{\max} \, \max(0,-a_2)
\end{cases}
\end{equation}
where $S_{\max}$, $F_{\max}$, and $B_{\max}$ denote the maximum steering angle, maximum engine force, and maximum braking force, respectively. In our configuration, the vehicle parameters are set as $S_{\max}=40^\circ$, $F_{\max}=800$, and $B_{\max}=150$, with a maximum attainable speed of 80 km/h.

\subsection{Implementation Details}
The proposed policy integrates a Transformer-based diffusion backbone with a sparse MoE module. The diffusion process follows DDPM \cite{ho2020denoising} with 100 denoising steps and a squared cosine noise schedule $\beta \in [10^{-4},0.02]$. Each forward pass generates an 8-step action horizon, from which the first step is executed before replanning. The MoE contains eight experts, with Top-$K$ routing ensuring sparse activation. Load-balancing and entropy regularization are used to prevent collapse and encourage expert diversity. Key hyperparameters are listed in Table~\ref{tab:parameter}. All experiments are conducted on a workstation with an Intel Core i7-14700K CPU, NVIDIA RTX 4080 SUPER GPU, and 32 GB RAM.

\begin{table}[htp]
    \centering
    \caption{Key hyperparameters of proposed model}
    \label{tab:parameter}
    \begin{tabular}{ccc}
    \toprule
    \textbf{Symbol} & \textbf{Meaning} & \textbf{Value} \\
    \midrule
    $H$ & Action horizon & 8 \\
    $B$ & Batch size & 64 \\
    $T$ & Number of diffusion steps & 100 \\ 
    $K$ & Number of active experts & 2 \\
    $N_{act}$ & Number of executed steps before replanning & 1 \\
    $N_{exp}$ & Number of experts in MoE & 8 \\
    $p_{drop}$ & Dropout rate in attention layers & 0.3 \\
    $\beta_{0}$ & Adam optimizer coefficients & 0.95 \\
    $\beta_{start}$ & Initial noise variance & $1\times 10^{-4}$ \\
    $\beta_{end}$ & Final noise variance & $0.02$ \\
    \bottomrule
    \end{tabular}
\end{table}

\subsection{Performance Evaluation}

\subsubsection{\textbf{Model scaling study}}
To assess the effect of model capacity on driving performance, we constructed four variants of the proposed policy by varying the embedding dimension, number of attention heads, and number of Transformer layers. These variants, denoted as \textit{Small}, \textit{Medium}, \textit{Large}, and \textit{Giant}, are summarized in Table~\ref{tab:model_size}, together with their total parameter counts (TP).

To assess the impact of model capacity on driving performance, we constructed four variants of the proposed policy by varying key architectural parameters including embedding dimension, number of attention heads, and number of Transformer layers. These variants, labeled as \textit{Small}, \textit{Medium}, \textit{Large}, and \textit{Giant}, are summarized in Table~\ref{tab:model_size}, along with their corresponding total parameter counts (TP).

\begin{table}[htp]
    \centering
    \caption{Key configurations of different model size}
    \begin{tabular}{ccccc}
    \toprule
         \textbf{Model} & \textbf{N emd} & \textbf{N head} & \textbf{N layer} & \textbf{TP(M)} \\
         \midrule
         Giant & 512 & 4 & 12 & 155.90 \\
         Large & 256 & 4 & 12 & 39.01 \\
         Medium & 256 & 4 & 8 & 26.26 \\
         Small & 128 & 2 & 4 & 3.39 \\
    \bottomrule
    \end{tabular}
    \label{tab:model_size}
\end{table}

\begin{figure}
    \centering
    \includegraphics[width=\linewidth]{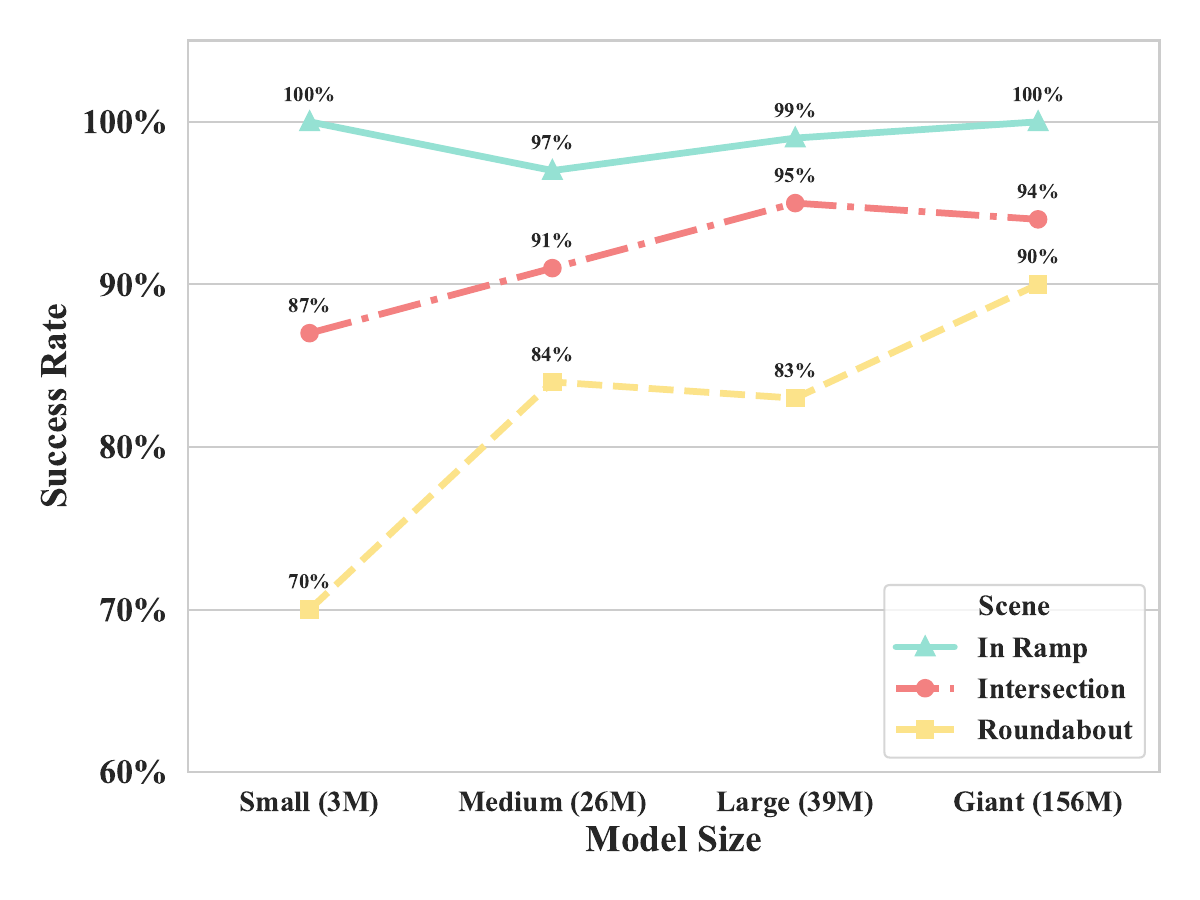}
    \caption{Model performance across different configurations, showing the relationship between model size and success rate in diverse driving scenarios.}
    \label{fig:model_size_success}
\end{figure}

As shown in Fig.~\ref{fig:model_size_success}, the overall trend indicates that larger models achieve higher success rates across the three representative driving scenarios. However, the improvement is not strictly monotonic. Although the \textit{Giant} model generally outperforms the smaller variants, intermediate models occasionally exhibit comparable or slightly higher performance in certain scenarios. This suggests that although increased capacity enhances the ability of the diffusion–Transformer backbone to capture multi-modal driving behaviors and maintain temporal consistency, the benefit saturates beyond a certain scale. Such non-monotonicity is consistent with observations in other generative models\cite{liu2023mtd}, where optimization stability and dataset coverage interact with model capacity in complex ways.

\begin{table}[htp]
    \centering
    \caption{Time consumption of different model sizes}
    \label{tab:decision_time}
    \begin{tabular}{c c c}
        \toprule
        \multirow{2}{*}{\textbf{Model}} & \multicolumn{2}{c}{\textbf{Time Consumption (ms)}} \\
        & \textbf{Avg.} & \textbf{Std.} \\
        \midrule
        Giant  & 81.61 & 0.47 \\
        Large  & 72.21 & 0.18 \\
        Medium & 50.41 & 0.24 \\
        Small  & 28.13 & 0.74 \\
        \bottomrule
    \end{tabular}
\end{table}

In terms of computational efficiency, decision-making time scales with model size, as shown in Table~\ref{tab:decision_time}. The \textit{Giant} model, with 155.9M parameters, achieves the highest success rate but incurs a decision latency of 81.6 ms. Despite this, it remains within the acceptable real-time threshold for driving systems. Consequently, the \textit{Giant} model was selected for subsequent experiments, as its performance benefits outweigh the marginal latency increase. In contrast, the \textit{Small} model offers faster inference at 28.1 ms but exhibits reduced robustness, particularly in more complex scenarios. The \textit{Medium} and \textit{Large} models provide a reasonable balance between performance and efficiency, though the \textit{Giant} model consistently delivers the best results across diverse driving environments, making it the preferred choice for tasks requiring higher reliability, such as intersection navigation and roundabout traversal.

\subsubsection{\textbf{Comparison with baseline models}}
To assess the performance of the proposed framework, we benchmark it against representative state-of-the-art methods spanning the three dominant paradigms of autonomous driving: reinforcement learning, rule-based decision-making, and imitation learning. Specifically, we employ PPO-Lag \cite{jayant2022model} as a safe reinforcement learning baseline, an enhanced Rotation Projection IDM (RPID) \cite{hu2019trajectory} as a rule-based baseline, and Implicit Behavior Cloning (IBC) with energy-based modeling \cite{florence2022implicit} as an imitation learning baseline. Together, these models provide a balanced and competitive foundation for evaluation.

\begin{table*}
    \centering
    \caption{Comparison of security, efficiency, and comfort test results of different methods in multiple scenarios.}
    \label{tab:Comparison}
    \begin{tabular}{llc ccc cc}
        \toprule
        \textbf{Scenario} & \textbf{Model} & \textbf{Success Rate} & \textbf{Collision Rate} & \parbox{2.5cm}{\centering \textbf{Average Episodic \\ Reward}} & \parbox{2cm}{\centering \textbf{Average \\ Velocity (m/s)}} & \parbox{2cm}{\centering \textbf{Acceleration \\ Variance}}& \parbox{2.5cm}{\centering \textbf{Average \\ Completion Steps}} \\
        \midrule
                    & PPO-Lag & 0.95 & 0.02 & 191.86 & 7.06 & 0.38 & 250.67 \\
        In Ramp     & RPID & 0.99 & 0.01 & 196.27 & 7.47 & 0.38 & 242.57 \\
                    & IBC & 0.86 & 0.11 & 190.28 & 8.03 & 0.35 & 220.78 \\
                    & Ours & \textbf{1.00} & \textbf{0.00} & \textbf{197.52} & \textbf{8.61} & 0.37 & \textbf{210.99} \\
        \midrule
                    & PPO-Lag & 0.90 & 0.10 & 116.79 & 6.30 & 0.45 & 172.02 \\
        Intersection& RPID & 0.63 & 0.37 & 98.57 & 7.28 & 0.39 & 125.42 \\
                    & IBC & 0.68 & 0.31 & 98.94 & 5.75 & 0.35 & 163.48 \\
                    & Ours & \textbf{0.94} & \textbf{0.06} & \textbf{121.54} & 6.34 & 0.45 & \textbf{174.05} \\
        \midrule
                    & PPO-Lag & 0.64 & 0.18 & 142.77 & 6.56 & 0.45 & 203.43 \\
        Roundabout  & RPID & 0.58 & 0.19 & 134.96 & 7.23 & 0.40 & 174.90 \\
                    & IBC & 0.70 & 0.22 & 139.71 & 6.41 & 0.35 & 245.98 \\
                    & Ours & \textbf{0.90} & \textbf{0.10} & \textbf{177.85} & 6.83 & 0.45 & 246.06 \\
        \bottomrule
    \end{tabular}
\end{table*}

We conduct comparisons across the three representative driving scenarios. The results, summarized in Table~\ref{tab:Comparison}, demonstrate the advantages of our knowledge-driven diffusion policy in terms of success rate, safety, and control efficiency.

In the In Ramp scenario, our model achieves a perfect success rate of 1.00 with zero collisions, outperforming all baselines, including PPO-Lag and IBC. This performance is accompanied by the highest episodic reward and average velocity, indicating both robust decision-making and smooth driving behavior. Notably, RPID performs well with a success rate of 0.99 but exhibits a slower average velocity, highlighting that while rule-based methods are safe in low-complexity environments, they tend to be overly conservative.

In the more challenging Intersection scenario, our model outperforms all baselines with a success rate of 0.94. In contrast, RPID shows a significant drop in success rate and a high collision rate, underscoring the limitations of rule-based methods in dynamic and interactive environments. In contrast, RPID shows a significant drop in success rate and a high collision rate, underscoring the limitations of rule-based methods in dynamic and interactive environments.

In the Roundabout scenario, which involves high interaction complexity, our model achieves a success rate of 0.90, significantly outperforming PPO-Lag and RPID, which struggle to manage multi-agent interactions. The episodic reward of our model is the highest among the approaches tested, coupled with a balanced acceleration variance of 0.45, indicating smooth and safe driving behavior. In contrast, IBC achieves a lower success rate, reflecting the challenges of purely data-driven methods in ensuring safety in dynamic environments.

The results demonstrate that our model outperforms the other approaches in dynamic driving environments, consistently achieving higher success rates and lower collision rates, particularly in complex scenarios such as intersections and roundabouts. While rule-based methods like RPID excel in simpler settings, they struggle with the varied and complex interactions present in real-world driving. Similarly, PPO-Lag provides safety, but its performance is limited by a lack of flexibility in handling multi-modal decision-making, as demonstrated in the Intersection and Roundabout scenarios. IBC, while effective in capturing multi-modal behavior, suffers from higher collision rates, particularly in the presence of dynamic obstacles.

\subsection{Ablation Study}
To evaluate the contribution of individual architectural components, we conducted an ablation study comparing three model variants: Ours, Baseline-T, and Baseline-U. Our model utilizes sparse activation with a Transformer backbone, while Baseline-T employs dense activation with the same Transformer backbone, and Baseline-U uses dense activation with a U-Net (CNN) backbone. The architectural details, total parameter counts, and Activation Parameters (AP) for each model are summarized in Table~\ref{tab:Model_Sparsity}.

\begin{table}[htp]
    \centering
    \caption{Architectural configurations of the models used in the ablation study}
    \label{tab:Model_Sparsity}
    \resizebox{0.5\textwidth}{!}{
    \begin{tabular}{ccccc}
         \toprule
         \textbf{Model} & \textbf{Sparsity} & \textbf{Temporal Backbone} & \textbf{TP(M)} & \textbf{AP(M)}  \\
         \midrule
         Ours & Sparse activation & Transformer & 155.90 & 56.24 \\
         Baseline-T & Dense activation & Transformer & 52.69 & 52.69 \\
         Baseline-U & Dense activation & U-Net(CNN) & 68.92 & 68.92 \\
         \bottomrule
    \end{tabular}
    }
\end{table}

\begin{figure}
    \centering
    \includegraphics[width=0.9\linewidth]{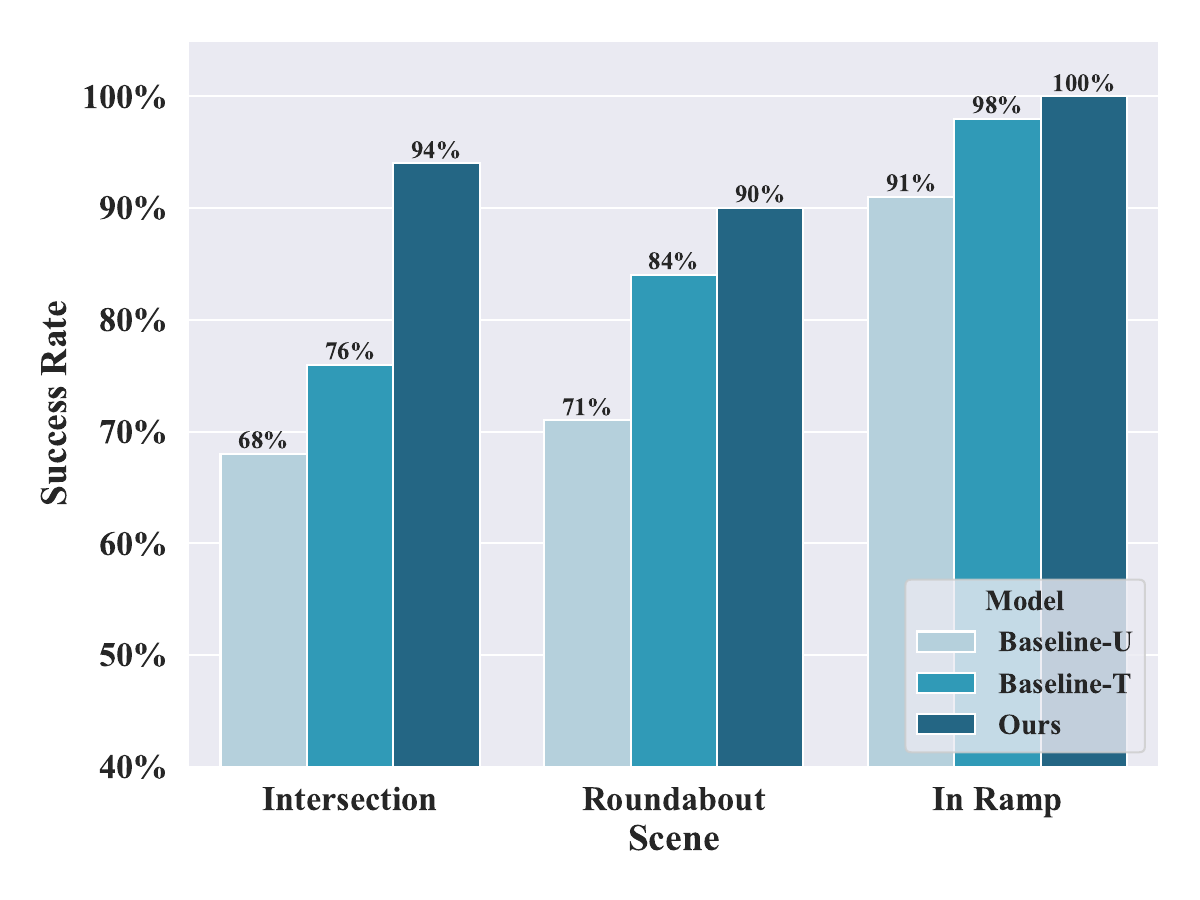}
    \caption{Performance comparison of model variants across driving scenarios.}
    \label{fig:ablation_figure}
\end{figure}

The performance results of these models are shown in Fig.~\ref{fig:ablation_figure}, where our model achieves the highest success rate, followed by Baseline-T, and the lowest is Baseline-U, which illustrates the necessity of the main components of the proposed model.

The introduction of the MoE mechanism with sparse activation facilitates a more efficient allocation of computational resources by activating only the most relevant parts of the model for a given input scenario. This leads to a lower AP but enhanced computational efficiency. While Baseline-T and Baseline-U have a higher parameter count, their performance improvements are not proportionate, particularly in complex scenarios like Intersection and Roundabout. This suggests that sparsity contributes to better resource management without compromising model performance.

The Transformer backbone used in Ours plays a crucial role in handling sequential decision-making. The Transformer excels at modeling temporal dependencies, which is essential for handling the long-term interactions and multi-modal decision-making required in autonomous driving. On the other hand, Baseline-U uses a U-Net backbone, which is more suited for tasks that require spatial understanding but struggles with temporal modeling and sequential decision-making, particularly in dynamic, multi-agent environments like Roundabout and Intersection. This explains the relatively poorer performance of Baseline-U in these complex scenarios.

The Transformer backbone in our model plays a critical role in sequential decision-making by effectively modeling temporal dependencies, which is essential for handling the long-term interactions and multi-modal decision-making required in autonomous driving. In comparison, Baseline-U excels in spatial understanding but struggles with temporal modeling and sequential decision-making. As a result, it performs poorly in dynamic, multi-agent environments, such as Roundabout and Intersection, where temporal consistency and long-term planning are vital.

This ablation study demonstrates that the MoE mechanism and Transformer backbone are pivotal to the performance of the proposed model. The MoE mechanism not only improves computational efficiency but also enhances the model's generalization capability by selecting expert knowledge units that are relevant to specific scenarios. The Transformer backbone is essential for maintaining temporal consistency and handling sequential decision-making. The results confirm that the combination of the MoE mechanism and Transformer-based temporal modeling provides a more robust and scalable solution for end-to-end autonomous driving tasks.

\subsection{Expert Activation Analysis}
To examine whether the MoE module captures reusable and scenario-specific knowledge, we analyze expert activation probabilities during closed-loop rollouts. Two complementary perspectives are considered: temporal activation patterns across time steps, and aggregated activation at the scenario level. Activation probability is defined as the router’s normalized weight assigned to each expert after Top-$K$ selection.

\begin{figure*}
    \centering
    \includegraphics[width=\linewidth]{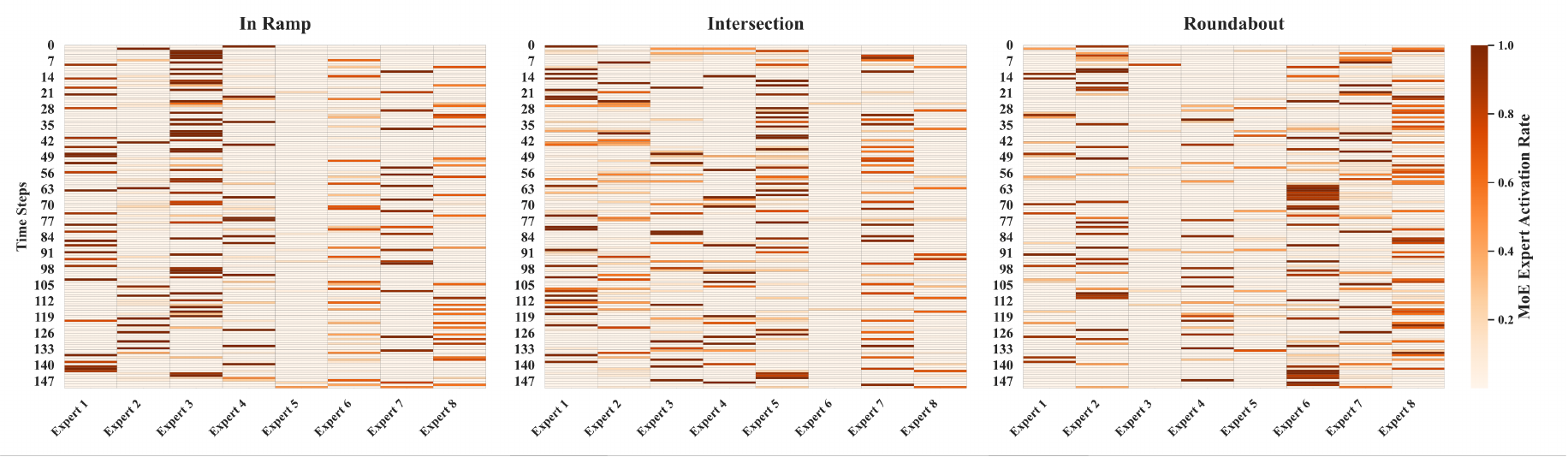}
    \caption{Temporal activation patterns of experts across driving scenarios. Expert activation exhibits a non-uniform temporal distribution. In the In Ramp, expert 3 is frequently activated. In the Intersection, expert 5 is mainly activated during the interaction. This characteristic is even more pronounced in the Roundabout, where expert 6 is concentratedly activated at the entrance and exit of the roundabout.}
    \label{fig:activation_time}
\end{figure*}

\subsubsection{Temporal activation patterns}
Figure~\ref{fig:activation_time} shows that activations are sparse and phase-dependent. Distinct bursts align with maneuver sub-phases rather than being uniformly distributed. In In Ramp, Expert 3 exhibits frequent low-variance activation throughout the episode, consistent with sustained longitudinal regulation and gap keeping during merging. In Intersection, Experts 1 and 5 show pronounced bursts at timesteps coinciding with vehicle–vehicle interactions inside the conflict zone, indicating engagement of interaction-sensitive competencies. In Roundabout, Experts 6 fire sharply at each entry and exit passage, while Expert 3 is largely quiescent, which is consistent with the reduced emphasis on car-following and the increased demand for precise lateral control. These temporally localized bursts indicate knowledge specialization that the router composes on demand as the scene evolves.

\begin{figure}
    \centering
    \includegraphics[width=\linewidth]{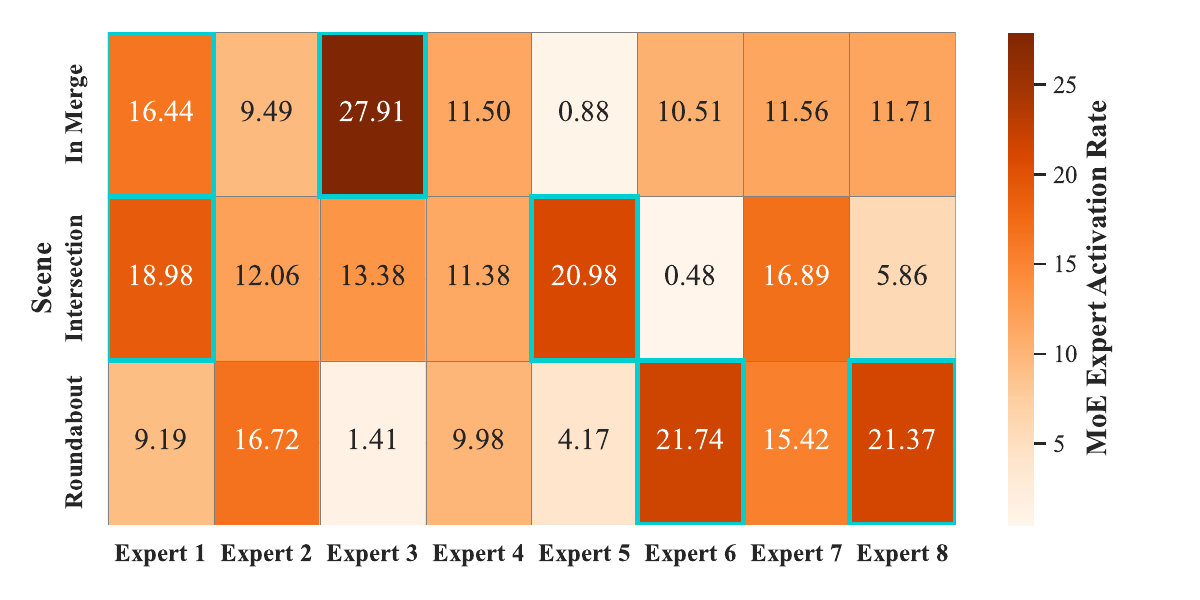}
    \caption{Scenario-level expert activation aggregated across episodes, where Experts 1 and 3 dominate in Merge, Experts 1 and 5 dominate in Intersection, and Experts 6 and 8 dominate in Roundabout, while Experts 2 and 4 exhibit occasional activation across multiple scenarios.}
    \label{fig:activation_scenario}
\end{figure}

\subsubsection{Scenario-level specialization and reuse} 
Figure~\ref{fig:activation_scenario} aggregates activation over full episodes and reveals a non-uniform and non-exclusive distribution across scenarios. Certain experts, such as Expert 3 in In Ramp, Expert 5 in Intersection, Experts 6 and 8 in Roundabout, dominate a single scenario, while others such as Experts 1 and 4 maintain non-trivial activation in multiple scenarios. The former suggests scenario-biased specialized knowledge like circular-flow negotiation in roundabouts, whereas the latter suggests reusable competencies just as lane keeping or longitudinal control that remain useful across contexts. Notably, In Ramp and Intersection both emphasize longitudinal regulation for the higher activation of Expert 1, but diverge in secondary demands where In Ramp favors following-distance control with higher Expert 3, while Intersection favors in-junction interactions with higher Expert 5. Roundabout shifts demand toward fine lateral control and gap acceptance at entries and exits, elevating Experts 6 and 8 and suppressing Expert 3.

Although we refrain from assigning fixed semantic labels to experts, the convergent regularities across Figs.~\ref{fig:activation_time} and \ref{fig:activation_scenario} support the  knowledge-driven nature. First, temporal sparsity indicates that experts are engaged as maneuver phases require, rather than being uniformly active. Second, cross-scenario reuse such as shared activation of Experts 1 and 4 coexists with scenario-biased specialization including Experts 3, 5, 6 and 8, implying a compositional policy that common competencies are reused broadly, while specialized ones are recruited when scene topology or interaction patterns demand them. This compositional routing provides a plausible mechanism for multi-scenario performance and generalization, that is, novel configurations can be addressed by recombining reusable experts with a small number of scenario-biased ones, without retraining a monolithic policy.

\subsection{Case Analysis}

\begin{figure*}
    \centering
    \includegraphics[width=\linewidth]{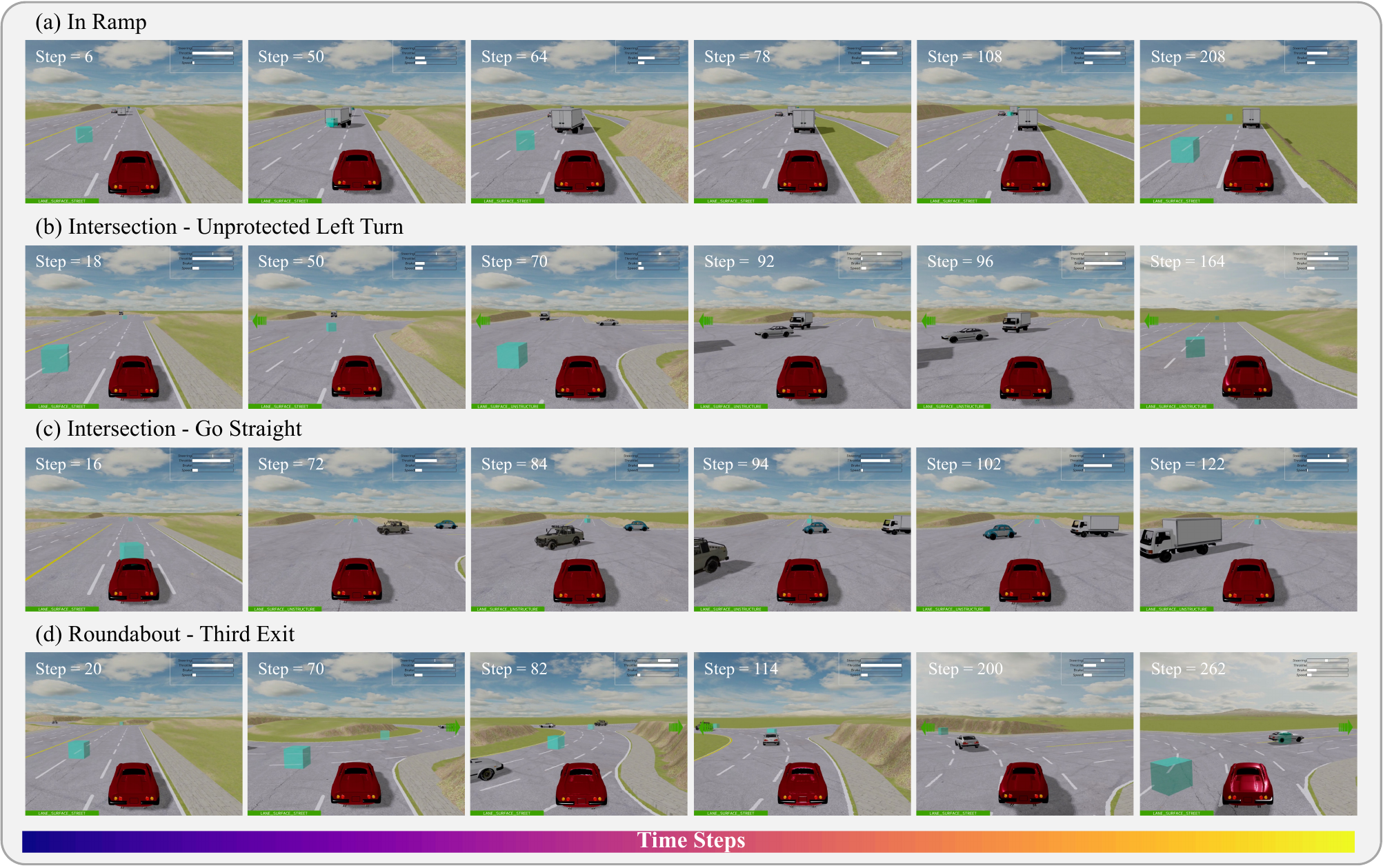}
    \caption{Representative case studies across different scenarios. (a) Ramp merging with cut-in, (b) unprotected left turn at an intersection, (c) intersection straight-through with interaction, and (d) roundabout navigation to the third exit. The ego vehicle demonstrates anticipatory control, smooth trajectory refinement, and adaptive interaction handling.}
    \label{fig:case_analysis}
\end{figure*}

To qualitatively illustrate the strengths of KDP, we present four representative closed-loop driving cases from distinct scenarios in Fig.~\ref{fig:case_analysis}). These examples highlight the model’s ability to adapt to diverse traffic interactions while ensuring both safety and efficiency.

\subsubsection{Ramp Merging with Cut-in} 
In the ramp merging case (Fig.~\ref{fig:case_analysis}a), the ego vehicle accelerates along the main lane and anticipates a potential cut-in from a truck in the adjacent lane. At Step 50, it executes a controlled deceleration to create a safe buffer, rather than reacting abruptly. After the truck stabilizes, the ego gradually resumes acceleration, maintaining a safe following distance until reaching the goal. This case highlights the model’s anticipatory capability and stable longitudinal control.

\subsubsection{Intersection: Unprotected Left Turn}
In the unprotected left-turn case (Fig.~\ref{fig:case_analysis}b), the ego vehicle approaches the junction as a vehicle from the right rapidly enters the conflict zone. At Step 92, the ego initiates deceleration and modulates braking intensity smoothly over several timesteps (Steps 92–96). This gradual adjustment illustrates how the diffusion-based policy generates smooth, comfortable trajectories instead of abrupt maneuvers. Once a safe gap emerges, the ego executes the turn efficiently, balancing caution with progress in a high-interaction setting.

\subsubsection{Intersection: Straight-through with Interaction}
The second intersection case (Fig.~\ref{fig:case_analysis}c) involves a cut-in maneuver by a leading vehicle near Step 72. The ego vehicle initially prepares to accelerate after the first vehicle passes, but adapts its decision as another vehicle accelerates in parallel. Ultimately, the ego yields and safely clears the intersection, which highlights the model’s flexibility in decision adjustment when facing uncertain multi-agent interactions.

\subsubsection{Roundabout: Navigation to the Third Exit}
In the roundabout case (Fig.~\ref{fig:case_analysis}d), the ego vehicle enters and navigates toward the third exit. At Step 82, a high-speed vehicle approaches from the rear-left. The ego responds by decelerating and adjusting its lateral position, maintaining precise angular control throughout. Before exiting, it reduces speed again to ensure safety when crossing another vehicle’s trajectory. This case underscores the model’s strength in fine-grained lateral control and safety assurance in complex circular flows.

\subsubsection{Multi-Scenario Evaluation}

\begin{figure}
    \centering
    \includegraphics[width=\linewidth]{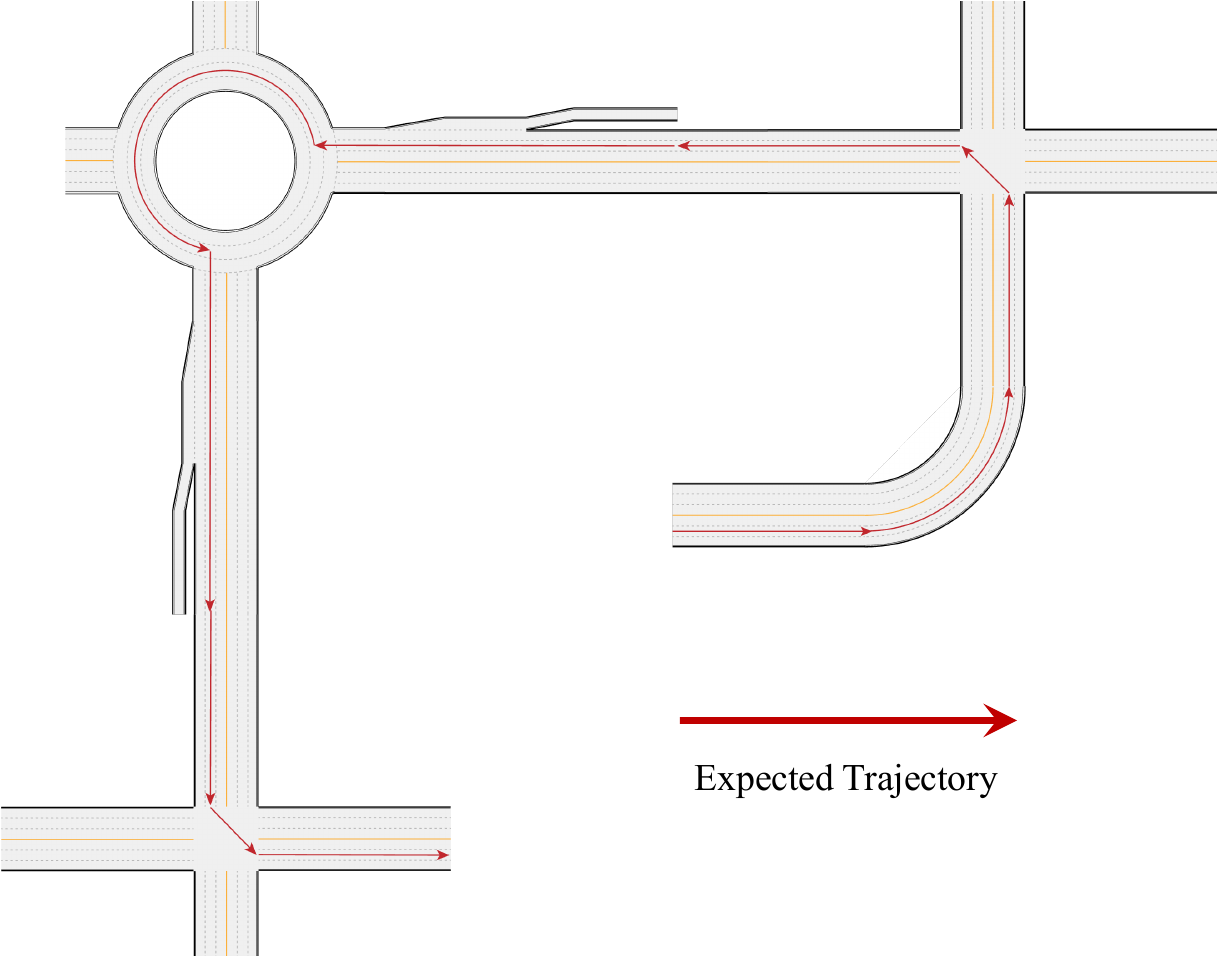}
    \caption{Overview of a large-scale continuous driving route used for multi-scenario evaluation. The red line indicates the trajectory followed by the ego vehicle.}
    \label{fig:multi_scenario}
\end{figure}

To further validate the global generalization and compositional adaptability of the proposed framework, we constructed large-scale continuous driving routes that incorporate both scenarios present in training and novel ones absent from the dataset.
These routes include curves, intersections, T-intersections, ramps, straights, and roundabouts, thereby requiring the policy to seamlessly transition across heterogeneous contexts within a single trajectory.
As illustrated in Fig.~\ref{fig:multi_scenario}, the ego vehicle successfully navigates diverse road structures and interaction patterns without task-specific tuning, demonstrating that the sparse expert routing mechanism can dynamically recombine reusable and specialized knowledge units to support robust performance in extended and realistic driving conditions.
A demonstration video is provided at this link\footnote{\href{https://perfectxu88.github.io/KDP-AD/\#casestudy}{Project page with additional video demonstration.}}.

Across these cases, the policy consistently exhibits anticipatory reactions, smooth trajectory refinement, and adaptive interaction handling. These demonstrations confirm that the integration of diffusion modeling with MoE-based routing enables knowledge-driven decision-making that generalizes effectively across diverse and challenging driving contexts.

\section{Conclusion}
This paper proposes KDP,  a knowledge-driven diffusion policy for end-to-end autonomous driving that integrates diffusion-based generative modeling with a sparse MoE routing mechanism. By modeling experts as abstract knowledge units, the framework achieves a balance between multi-modal expressiveness, temporal stability, and modular generalization. Experimental results across representative driving scenarios demonstrate superior performance over baselines, while ablation and activation analyses confirm the critical role of sparse expert routing and reveal structured patterns of specialization and reuse. Case studies further illustrate the policy’s capacity for anticipatory, safe, and adaptive decision-making in complex interactive environments.
Although the present evaluation is limited to simulation and selected scenarios, future research will focus on extending the approach to real-world datasets, incorporating richer sensory modalities, and developing continual learning mechanisms to handle open-world conditions.

\ifCLASSOPTIONcaptionsoff
  \newpage
\fi

\bibliographystyle{unsrt}
\bibliography{related}

\end{document}